\documentclass[lettersize,journal]{IEEEtran}
\usepackage{amsmath,amsfonts}
\usepackage{algorithmic}
\usepackage{algorithm}
\usepackage{array}
\usepackage[caption=false,font=normalsize,labelfont=sf,textfont=sf]{subfig}
\usepackage{textcomp}
\usepackage{stfloats}
\usepackage{url}
\usepackage{verbatim}
\usepackage{graphicx}
\usepackage{cite}
\usepackage{amsmath}
\usepackage{amssymb}
\usepackage{ragged2e} 
\usepackage{booktabs,makecell, multirow, tabularx}
\usepackage{xcolor}

\begin{document}

\title{T2IW: Joint Text to Image \& Watermark Generation}

\author{An-An Liu$^\#$, Guokai Zhang$^\#$, Yuting Su, Ning Xu, Yongdong Zhang, and Lanjun Wang$^*$
\thanks{An-An Liu, Guokai Zhang, Yuting Su, Ning Xu, and Lanjun Wang are with Tianjin University, Tianjin 300072, China (email: wanglanjun@tju.edu.cn).}
\thanks{Yongdong Zhang is with University of Science and Technology of China, Hefei 230026, China.}
\thanks{$\#$ These authors contributed equally to the work.}
\thanks{$*$ Corresponding author, to whom correspondence should be addressed.}}

\markboth{Arxiv Preprint}%
{Shell \MakeLowercase{\textit{et al.}}: A Sample Article Using IEEEtran.cls for IEEE Journals}


    \maketitle

\begin{abstract}
Recent developments in text-conditioned image generative models have revolutionized the production of realistic results. Unfortunately, this has also led to an increase in privacy violations and the spread of false information, which requires the need for traceability, privacy protection, and other security measures. However, existing text-to-image paradigms lack the technical capabilities to link traceable messages with image generation. In this study, we introduce a novel task for the joint generation of text to image and watermark (T2IW). This T2IW scheme ensures minimal damage to image quality when generating a compound image by forcing the semantic feature and the watermark signal to be compatible in pixels. Additionally, by utilizing principles from Shannon information theory and non-cooperative game theory, we are able to separate the revealed image and the revealed watermark from the compound image. Furthermore, we strengthen the watermark robustness of our approach by subjecting the compound image to various post-processing attacks, with minimal pixel distortion observed in the revealed watermark. Extensive experiments have demonstrated remarkable achievements in image quality, watermark invisibility, and watermark robustness, supported by our proposed set of evaluation metrics.
\end{abstract}


\section{Introduction} 
\IEEEPARstart{T}{he} demand for customized and innovative creation in the realm of art, marketing, and entertainment sectors continues to grow prosperous, inducing artificial intelligent generated content (AIGC) platforms emerging in large numbers~\cite{Tao00JBX22,ye2023recurrent,ZhangS21,ZhangXL17}. Until recently, text-to-image (T2I) generative models can depict paintings that rival professional photography and masterpiece artworks~\cite{HoJA20,RombachBLEO22,Tao_2023_CVPR}. The plausibility of synthetic images has beyond the scope of human capability to discern authenticity, resulting in frequent incidents such as privacy breaches and rumor forgery. Sadly, it has led to notable legal conflicts that Stability AI and Midjourney are embroiled in a lawsuit together~\cite{Strowel23}. In order to safeguard security, multiple technology giants, including OpenAI and Google, have issued a statement that safe, secure, and truth-worthy AI will be developed by integrating invisible and identifiable watermarks into the generated images in their future products~\cite{wu2023comprehensive}. Furthermore, it can be anticipated that policymakers will put more rational regulatory measures in place for T2I generation.

\begin{figure}\centering
	\centering
	\includegraphics[scale=0.82]{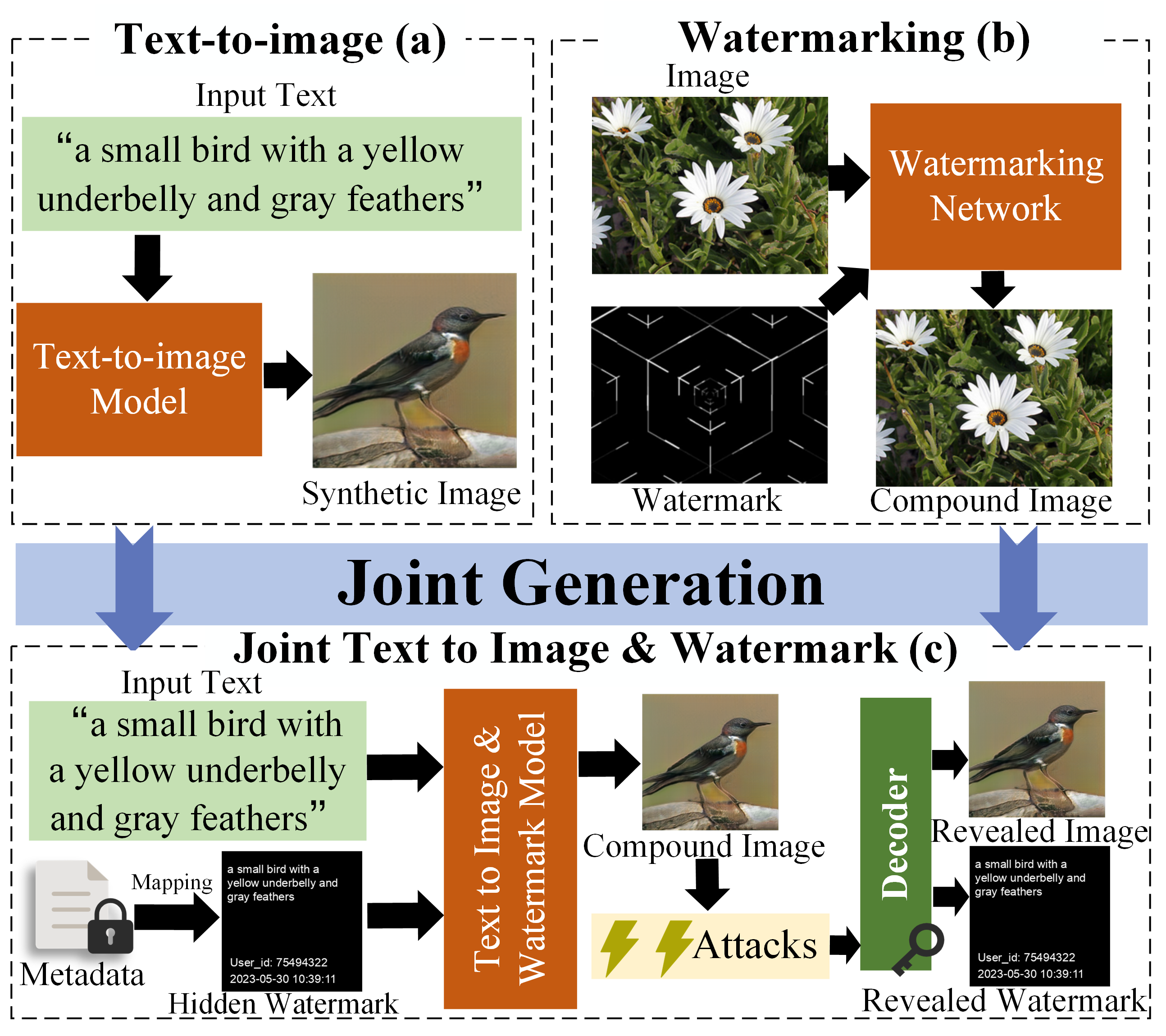}
	\vspace{-1ex}
	\caption{Workflow of text-to-image, watermarking, and our proposed T2IW. \textbf{(a) Text-to-image} synthesizes a high-quality image conditioned on the input text. \textbf{(b) Watermarking} hides a watermark into a real image invisibly. \textbf{(c) Text to image \& watermark} seeks to integrate a message-bearing watermark into the image generation procedure, thereby enabling the generation of the compound image and the decoding of the revealed watermark and the revealed image, even in the presence of attacks.}  
	\vspace{-1ex}
	\label{fig:workflow}
\end{figure}

Adding watermarks on the images is a classical task. 
However, applying traditional methods of generating images first and then adding watermarks~\cite{ZhuKJF18,DingMCL22}, on the T2I task poses the potential risk of information tampering between consecutive stages (which are the image generation stage and the watermarking stage).
Moreover, for the T2I task, these methods ignore information from the text part or other corresponding metadata of the generation scenario, e.g., creation time stamp, the name of the creator who inputs the text query, etc.  
Meanwhile, such information (i.e., both text and metadata) is very important for the traceability of image generation, which helps to differentiate legal responsibility between the creator and the generative model when the output image is harmful. As a result, we develop a new AIGC task to generate the image and the watermark jointly from the text, i.e., text to image \& watermark (T2IW), the workflow of which is shown in Fig.~\ref{fig:workflow} (c). More specifically, T2IW is to generate an image that is accompanied by an invisible watermark, which can synchronously record metadata, avoiding the risk of information tampering in traditional watermarking schemes.
Currently, there have been no attempts to create an image with a watermark from text together.
Firstly, it is not easy to obtain a good visual quality in the output image. This is attributed to two main challenges. One is that the placement of a watermark has shown to be difficult to balance between effective concealment and preservation of the overall appearance of the generated image~\cite{SuWJZLL15}. The other is that previous T2I techniques~\cite{Tao00JBX22,ye2023recurrent,ZhangS21,ZhangXL17} have not been able to effectively map text to image pixels. Therefore, in this study, we need T2IW to produce both invisible watermark and visuals comparable to the T2I methods.
Secondly, the output compound image (i.e., the generated image with a watermark) must be powerful enough to recover the watermark even when exposed to a variety of attacks that are designed to destroy it. Unfortunately, some existing techniques~\cite{WangD19,ZhangLJYXZ19} are susceptible to distortion, resulting in the loss of essential watermark information. Therefore, this T2IW study requires the production of robust compound images against a broad range of attacks with varying levels of intensity.
Finally, traditional approaches focus on the incorporation of a single type of watermark (e.g., a logo or an identifiable visual pattern) into an image~\cite{HasanICKA21,ZearSK18,PalR18,7754176,SinghS17a,AbdulrahmanO19}. However, in the age of AI-generated content, multiple pieces of information must be included in the generated image to trace its origin, such as the creator's name and the time of creation. To meet this need, T2IW is designed to incorporate multiple sources of information to watermark the image.


In this study, we propose a T2IW framework for hiding watermarks in generated images, adapting to T2I models. As shown in Fig.~\ref{fig:pipeline}, the framework has three components: \textit{joint generation} for synthesizing the compound image, \textit{image decoupling} for separating the revealed image and the revealed watermark, and \textit{optimization strategy} for setting objective functions of the compound image, the revealed image and the revealed watermark. 
In detail, we use the U-Net~\cite{RonnebergerFB15} to improve the visual quality by combining the watermark and the image. We adjust the frequency of semantic feature fusion to reduce the interference of the hidden watermark while creating the compound image.
Furthermore, we improve the non-cooperative game~\cite{ritzberger2002foundations} in the decoupling of a compound image to develop an allocation policy for extracting the revealed watermark and the revealed image. This requires the watermark decoder to be trained to respond to various attacks, thus enhancing the robustness of the watermark. It is also important to note that each generated image is linked to a unique watermark that incorporates multiple sources of information, enabling both embedding and reconstruction processes. Additionally, we design a set of evaluation metrics for the T2IW task, which is evaluated from the perspectives of image quality, watermark invisibility, and watermark robustness. These metrics will be thoroughly explained in Sec.~\ref{metrics}. 

The main contributions of this paper can be summarized into five-folds as follows:
\begin{itemize}
	\item We introduce a new problem by jointing T2I and watermarking, which integrates diverse message-carried watermarks into generating images for traceability, privacy preservation and fulfilling other security goals.
	\item We propose a joint approach for T2IW that is suitable for embedding the watermark signal into the generated image without being visible. 
 Furthermore, Shannon information theory and non-cooperative game theory are used to separate the revealed watermark and the revealed image, allowing for a compromise in the information distribution of the compound image.
	\item A customized evaluation scheme has been developed for the T2IW, evaluating from the perspectives of image quality, watermark invisibility and watermark robustness.
	\item We employ a data augmentation strategy to actively apply post-processing attacks to the compound image in order to obtain robust parameters for the watermark decoder. This strategy ensures that the watermark is highly resistant to common attacks of varying intensities.
    \item Extensive experiments on two categories of models (i.e. single-stage and multi-stage generation) have revealed that our proposed framework does not significantly reduce the visual quality of the compound image with minimal disruption. Moreover, after external post-processing attacks (e.g., rotation, random cropping, Gaussian noise, salt and pepper noise, brightness, Gaussian blur, etc.), the messages in the watermark can still be maintained and recovered.
\end{itemize}

\section{Related Work}
\label{related_work}
In this section, we present previous research conducted in two distinct fields: text-to-image generation and image watermarking.
\subsection{Text-to-Image Generation}
Until recently, visually realistic images are mostly produced from texts by generative adversarial networks (GANs)~\cite{Tao00JBX22,ye2023recurrent,ZhangS21,ZhangXL17} and denoising diffusion probabilistic models (DDPMs)~\cite{HoJA20,RombachBLEO22}. The literature~\cite{Tao_2023_CVPR} has shown that lightweight GAN architecture still has the ability to compete with DDPM. To verify the effectiveness and generalization of our proposed scheme, we employ the classical GAN structures in this paper and divide them into two categories, named single-stage generation and multi-stage generation.

Firstly, single-stage GANs~\cite{ye2023recurrent,Tao00JBX22,Tao_2023_CVPR} briefly encode the text description to generate images, without any intermediate representations. Multi-stage GANs~\cite{ZhangXL17,ZhangXLZWHM19,XuZHZGH018} regularly stack three combinations of generator-discriminators, forcing growth in resolution with depicting details. Nevertheless, the above mentioned networks lack of message hiding interface, unable to cope with balancing the appearance of the watermark signal and the text semantic. Then, the research status of two types of methods will be introduced in detail below.

\subsubsection{Single-stage Generation} This paradigm usually uses one generator and one discriminator to synthesize images from text efficiently. Initially, Reed et al.~\cite{ReedAYLSL16} first used CNN-based GAN to generate images from linguistic descriptions encoded by RNN. Furthermore, Dong et al.~\cite{DongYWG17} proposed an end-to-end generator composed of residual blocks to map the text to the corresponding area of the reference image. Since GANs incur mode collapse, Cha et al.~\cite{ChaGK19} alleviated the problem by forming positive and negative training examples, ensuring the diversity of the generated images. Before Tao et al.~\cite{Tao00JBX22}. proposed DF-GAN, the single-step paradigm was criticized in synthesizing high-quality images. They explicitly deepened the integration of text-image features, directly obtaining the photo-realistic images. Inspired by DF-GAN~\cite{Tao00JBX22}, Zhang et al.~\cite{ZhangS21} linearly stacked the convolutional structure to weaken the training cost. In addition, Ye et al.~\cite{ye2023recurrent}, built on~\cite{Tao00JBX22}, modeled the long-term dependency between affine transformation blocks with significant performance improvement. 

\subsubsection{Multi-stage Generation} The superiority of the approaches that perform multi-stage~\cite{ZhangXL17,ZhangXLZWHM19,XuZHZGH018} is to sculpture details from low to high resolution gradually. From this Zhang et al.~\cite{ZhangXL17} realized that the image distribution based on roughly calibrated low-resolution images was more likely to overlap with the real image distribution and proposed an end-to-end network called StackGAN. Technically, conditioning augmentation was designed to smooth the text feature space to increase diversity, generalizing in subsequent methods~\cite{ZhangXLZWHM19,XuZHZGH018}. Later, they proposed a function-enhanced version named StackGAN++~\cite{ZhangXLZWHM19} to unified conditional and unconditional T2I.  Moreover, Xu et al.~\cite{XuZHZGH018} introduced the attention mechanism~\cite{VaswaniSPUJGKP17}, inherited from StackGAN++~\cite{ZhangXLZWHM19}, to discover which region needs to be refined. In addition, there were numerous methods to introduce auxiliary representations to hierarchically help generate high-resolution images. SD-GAN~\cite{YinLSYWS19} used the Siamese structure~\cite{SalimansGZCRCC16,HeuselRUNH17} to distill the semantic commons of texts for consistency in image generation. Peng et al.~\cite{PengZSCWHJ22} used exogenous knowledge to imitate the behavior of human painting. Notice that the layout can be used to draw patches on a canvas. To control what and where to generate, Reed et al.~\cite{ReedAMTSL16} first used bounding box and keypoint as intermediate representations to guide the placement of multiple objects. Taking into account the dilemma in capturing the dependencies of multiple objects, there were many attempts to effectively combine them at the object level~\cite{WuLHHC22,HinzHW19,LiZZHHLG19, ZhaoMYS19}. Most prominently, OP-GAN~\cite{HinzHW22} enhanced the control of words' layout in images, demonstrating the powerful capability to generate high-resolution images.  Worth mentioning,~\cite{Tingting213,ZhuP0019,YinLSYWS19} also raised mutli-stage GANs' capacity.

To verify the general applicability of our framework, we select two representative models, RAT-GAN~\cite{ye2023recurrent} and AttnGAN~\cite{XuZHZGH018}, which incorporate the characteristic mentioned above.

\subsection{Image watermarking}
Image watermarking~\cite{WanWZLYS22} aims to embed secret messages into general digital image carriers, to be received by the consent holder and not hacked by attackers. Guarantee that the public media carriers suffer minimal damage and ciphertext seeks to be robust and imperceptible. By virtue of the application properties, the target images of watermarking can be categorized into real images~\cite{HasanICKA21,7754176,DingMCL22} and generated images~\cite{WuLYZ21,OngCNFY21,YuSAF21}, both of which have technical similarities and will be unfolded for narration in this section.


\subsubsection{Watermarking on Real Images} The techniques can be divided into two main categories, which refer to the spatial domain through the discrete cosine transform (DCT)~\cite{HasanICKA21,ZearSK18,PalR18} and the discrete wavelet transform (DWT)~\cite{7754176,SinghS17a,AbdulrahmanO19}. The powerful learning capabilities of deep neural networks, similar to those found in steganography~\cite{KadhimPVH19}, gradually become apparent. Kandi et al.~\cite{KandiMS17} first adopted convolutional neural network (CNN) in image watermarking. Discovered by~\cite{ZhuKJF18}, neural networks could learn to use invisible perturbations to encode a binary message into an image. Then, Zhang et al.~\cite{03892} introduced a groundbreaking method that uses generative adversarial networks to conceal arbitrary binary data within images. This technique empowered GAN to enhance the visual appeal of images while effectively hiding embedded information. A remarkably impressive work~\cite{TancikMN20} introduced the learned steganographic algorithm to enable robust encoding and decoding of arbitrary hyperlink bit-strings into photos in a manner that approaches perceptual invisibility. Recently, Ding et al.~\cite{DingMCL22} proposed a simple and effective deep neural network (DNN) to fuse the image and the watermark for better generalization, from the perspective of intellectual content protection legislation. In conclusion, deep learning-based image watermarking technology is still in its infancy and will gain more attention.

\subsubsection{Watermarking on Generated Images} In the AIGC era, with the recent enhancement of the powerful generation capacity of generative models~\cite{Tao00JBX22,ye2023recurrent,ZhangS21,ZhangXL17}, there has been a surge in generated content. As a result, researchers have gradually started to explore the technology of embedding watermarks in generated images. To protect the intellectual property of generative adversarial networks, Wu et al.~\cite{WuLYZ21} encoded a binary sequence in the generated image. Then, a pre-trained CNN watermarking decoding block was inserted to predict the hidden sequence for ownership verification. In essence, it still performed adding messages into the determined image, not participating in the process of image generation. Purposefully, in common black-box scenarios, Yu et al.~\cite{OngCNFY21} proposed a backdoor-based framework to set triggers on inputs. However, the watermark was visible, and the generative model needed specific inputs as queries. By building the watermarked training data in~\cite{YuSAF21}, Yu et al. trained a model to automatically output watermarked images. The inherent drawback of creating watermarked datasets was the significant cost involved in the manufacturing process. Since there is currently no watermarking technology specifically designed for the T2I generation task, we decide to delve into the joint generation of both the image and the watermark from text.

\begin{figure*}[!t]
	\centering
	\includegraphics[width=\textwidth,scale=0.75]{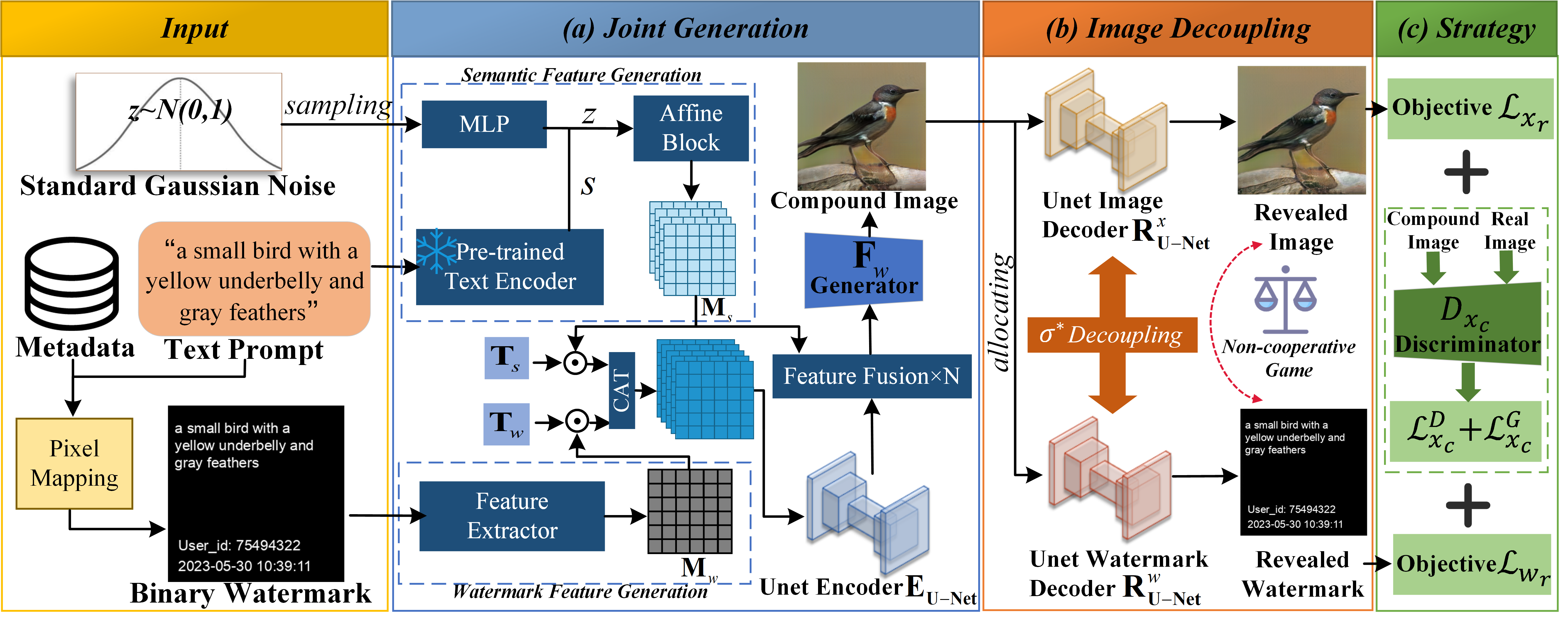}	
	\vspace{-5ex}
	\caption{Overview of our proposed T2IW framework, compromising three main components. (a) The purpose of joint generation is to incorporate noise, text, and watermark signals to create a compound image, which essentially means generating an image with the invisible watermark. (b) Image decoupling utilizes the non-cooperative game theory to establish a pair of decoders and the information allocating strategies, enabling the decoupling of the image and watermark from a compound image. (c) The optimization strategy encompasses the objective functions for the revealed image, the revealed watermark and the compound image.}
	\vspace{-3ex}
	\label{fig:pipeline}
\end{figure*}

\section{Problem Statement}
\label{problem_statement}
The primary objective of T2IW is to create the image embedded with the watermark (that is, the compound image), with the intention of subsequently separating them into the revealed watermark and the revealed image. Formally, let $i_t$, $i_z$ and $i_m$ denote the input text prompt, the standard Gaussian noise sampling, and the creation-related metadata, respectively. The hidden binary watermark is encoded as $w_h={\delta_w(i_t,i_m)}$, where $\delta_w(\cdot)$ represents a function that maps characters to pixels to produce the watermark. Given the aforementioned components, T2IW facilitates both watermark embedding and recovery from the compound image, which process is outlined as follows:
\begin{equation}
	\label{task_define}
	\begin{split}
		x_{c} &= \boldsymbol{\Psi}(i_t,i_z,w_h;\Theta_{G})\\
		x_{r}, w_{r} &= \boldsymbol{\Omega}(x_{c};\Theta_{D}) 
	\end{split}
\end{equation}
where the output compound image is represented as $x_{c}$, while the revealed image and the revealed watermark is represented as $x_{r}$ and $w_{r}$, respectively. The functions $\boldsymbol{\Psi}(\cdot)$ and $\boldsymbol{\Omega}(\cdot)$ represent the compound image generation and decoupling process, with the set of learnable parameters $\Theta_{G}$ and $\Theta_{D}$ configured. 

In accordance with Eq.~\ref{task_define}, distinct properties are assigned to the three outputs from T2IW:
\begin{itemize}
	\item \textit{Property for $x_c$:} The compound image aims to exhibit a visually appealing effect, comparable to the quality of image generated by corresponding T2I method. Afterward, the hidden watermark can be reconstructed using a specific decoder from it.
    \item \textit{Property for $x_r$:} The revealed image tries to closely resemble $x_c$ in terms of visual appearance, while the watermark cannot be parsed from it.
    \item \textit{Property for $w_r$:} The main emphasis of the revealed watermark lies in achieving high-quality reconstruction and strong robustness, particularly in making the messages attached to the watermark recognizable.
\end{itemize}

\section{Methodology}
\label{methodology}
In this section, we first outline our proposed T2IW framework. Following that, we divide the framework into three distinct components and offer a detailed derivation. 

\subsection{Framework}
The overview of our proposed T2IW is presented in Fig.~\ref{fig:pipeline}. Accordingly, we introduce three phases: joint generation in Fig.~\ref{fig:pipeline} (a), image decoupling in Fig.~\ref{fig:pipeline} (b), and optimization strategy in Fig.~\ref{fig:pipeline} (c), adapting to various generative adversarial architectures~\cite{Tao00JBX22,ye2023recurrent,ZhangS21,ZhangXL17}. Consequently, joint generation is to generate a compound image, equal to the function $\boldsymbol{\Psi}(\cdot)$ in Eq.~\ref{task_define}, while image decoupling is to extract the revealed image and the revealed watermark, equal to the function $\boldsymbol{\Omega}(\cdot)$. Eventually, the optimization strategy stage provides objective functions for the compound image, the revealed image, and the revealed watermark, respectively. In summary, T2IW provides new insight into the traceability of generated content by jointly generating the image with the watermark from text.

\subsection{Joint Generation}
The goal of joint generation is to integrate the binary watermark into the feature domain of the generated image invisibly, thereby ensuring the compatibility of textual semantic and watermark signal in pixels. Following the setting in the previous studies~\cite{Tao00JBX22,ye2023recurrent}, we use a pre-trained bidirectional LSTM as a text encoder to encode the text description into a sentence-level feature vector $s\in\mathbb{R}^{L\times1}$ and a learnable MLP to embed the noise sampling from standard Gaussian distribution $\mathcal{N}(0,1)$ into the noise vector $z\in\mathbb{R}^{L\times1}$. To enrich visual expression, the affine transformation~\cite{Tao00JBX22}, i.e. $\mathrm{Affine}(z,s) = \mathrm{\gamma}_{scale}(s)\cdot \eta(z)+\mathrm{\gamma}_{shift}(s)$, is employed, where $\mathrm{\gamma}_{scale}(\cdot)$, $\mathrm{\gamma}_{shift}(\cdot)$, and $\eta(\cdot)$ represent the scaling, the shifting, and the noise mapping functions. The output matrix of the affine transformation is represented as $\mathbf{M}_s \in \mathbb{R}^{H\times W\times C}$, considered a semantic feature.

To prohibit serialization errors whilst decoding and dramatically downscale the perturbing in image generation (e.g. sequence encoding)~\cite{2014Robust}, we have noticed the binary-form image has the traits to retain the secret messages with extensible capacity. Then, we embed the text prompt, 8-digit user identity (randomly sampled from a uniform distribution $u_{id}\sim\mathbb{U}(0,9)$ in our implementation), and the creation time stamp as the messages, which possess a record of creative elements for certain tracking capability. Subsequently, these messages are assigned to binary pixels as the hidden watermark $w_h$. Finally, the CNN is adopted as a feature extractor to obtain the watermark feature, denoted $\mathbf{M}_w \in \mathbb{R}^{H\times W\times 1}$.

To augment the expressive capacity of the textual semantic and optimize the concealment of the watermark, it is crucial to seek a co-embedded feature space for integrating the $\mathbf{M}_s$ and the $\mathbf{M}_w$ meanwhile. The learnable matrices $\mathbf{T}_s$ and $\mathbf{T}_w$ are set to align the semantic and the watermark feature sizes, and the formulation is as follows:
\begin{equation}
	\label{fusion_watermark}
	\begin{split}
	\mathbf{M}_i^{c}=
	\begin{cases}
		\mathbf{E_{\mathrm{U-Net}}}([\mathbf{T}_s \mathbf{M}_s, \mathbf{T}_w\mathbf{M}_w]), i=1\\
		\mathbf{Y}_i(\mathbf{M}_{i-1}^{c},\mathbf{M}_{s};\theta_i), i=2,3,...,N
	\end{cases}
	\end{split}
\end{equation}
where $\mathbf{Y}_i(\cdot)$ is the $i$-$th$ feature fusion block, mainly composed of multi-layer convolution with nearest interpolation, $\mathbf{M}_i^{c}$ is the $i$-$th$ visual feature map, $N$ denotes the feature coupling iteration number, $\mathbf{E_{\mathrm{U-Net}}}(\cdot)$ refers to a U-Net as the encoder to couple the semantic feature with the watermark feature, $[\cdot,\cdot]$ means the channel-wise concatenation. Finally, the generator plays a vital role in generating the compound image as follows:
\begin{equation}
	\label{generate_img}
	 x_{c} = \boldsymbol{\mathbf{F}_w}(\mathbf{M}_{N}^{c};\Theta _\epsilon)
\end{equation}
where $x_c$ refers to the compound image, which exhibits excellent imperceptibility of the embedded watermark, as shown in Fig.~\ref{fig:pipeline} (a). $\boldsymbol{\mathbf{F}_w}(\cdot)$ denotes the regular generator parameterized by $\Theta _\epsilon$, which can be transplanted from existing methods~\cite{ye2023recurrent,ZhangXLZWHM19}.

\begin{figure}\centering
	\centering
	\includegraphics[scale=0.70]{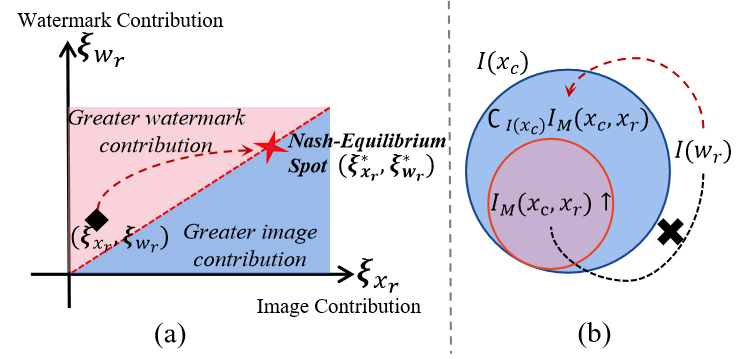}
	\vspace{-1ex}
	\caption{Illustration of the non-cooperative game on the T2IW scenario. (a) represents the schematic diagram of arbitrary spot $(\xi_{x_r},\xi_{w_r})$ on the hyperplane approaching the Nash equilibrium $(\xi^*_{x_r},\xi^*_{w_r})$, and (b) represents the strategy implementation during watermark revealing.}
	\vspace{-1ex}
	\label{fig:nash}
\end{figure}

\subsection{Image Decoupling}\label{decouple} 

Given that the compound image can be obtained using Eq.~\ref{generate_img}, the image decoupling process is to find the optimal information allocation policy for decoupling the image and the watermark, both of which individually contribute to the overall appearance of the compound image to different extents. Suppose that there are contribution factors $\xi_{x_r}$, $\xi_{w_r}$ and $\xi_{x_c}$ for reflecting visual effects from $x_r$, $w_r$ and $x_c$ respectively, approximated as the allocation strategy with a linear positive correlation:
\begin{equation}
	\label{allocation}
	\begin{split}
		\xi_{x_c}\propto  (1-\sigma(x_c))\cdot \xi_{x_r} + \sigma(x_c)\cdot \xi_{w_r}
	\end{split}
\end{equation}
where the $\sigma(\cdot)$ reflects the allocation strategy for distributing information of the compound image $x_c$ to the revealed image $x_r$ and the revealed watermark $w_r$. This allocation policy can be viewed as a non-cooperative game~\cite{ritzberger2002foundations}, characterized by the absence of communication and negotiation among players, requiring them to devise their dominant strategies~\cite{ritzberger2002foundations}. Specifically, the players place significant emphasis on autonomous decision-making to maximize their individual interests, independent of the strategies employed by others within the strategic environment. Purposefully, there is a trade-off for the benefit of both players to be approximated, also termed as Nash equilibrium~\cite{ritzberger2002foundations}. For game $G=\{s_1,...,s_n;p_1,...p_n\}$, assume that $(s_1',...,s_n')$ is the arbitrary strategy collection, the strategy $s_i^*$ is the optimal choice for player $p_i$ when encountering the combination of strategies $(s_1',...,s_{i-1}',s_{i+1}',...,s_n')$ of other players. Formally, the non-cooperative game can be denoted as:
\begin{equation}
	\label{game}\small
	\begin{split}
		\sigma^* =(s_1^*,...,s_n^*)=  &\mathop{\mathrm{argmax}}\limits_{(s_1,...,s_n)} \mathrm{\Phi}_{gain}^{P}(s_1,...,s_n)\\
		\mathrm{\Phi}_{gain}^{p_i}(s_1',...,s_i^*,...,s_n')&\geq \mathrm{\Phi}_{gain}^{p_i}(s_1',...,s_i',...,s_n'),\forall i \in [1,n]
	\end{split}
\end{equation}
where $\sigma^*$ denotes the Nash equilibrium strategy collection, where no player can increase their profits by changing their strategies individually, ${P}$ is a set of players, $\mathrm{\Phi}_{gain}^{p_i}(\cdot)$ denotes the gains by executing strategy collection for player $p_i$, while $\mathrm{\Phi}_{gain}^{P}(\cdot)$ denotes the total gains for all players $P$. 

In the joint T2IW scenario, $x_r$ and $w_r$ engage in a non-cooperative game and develop selfish strategies towards Nash equilibrium. From an intuitive point of view, the hyperplane depicted in Fig.~\ref{fig:nash} (a) indicates that there is one Nash equilibrium spot to impel the appearance of the prime image. Empirically, the spot is more biased towards $x_r$, which needs to maintain consistency with $x_c$ as much as possible. We then transfer the contribution factors and assume that $\xi_{x_c}$ reaches the fixed value $\xi^*_{x_c}$, indicating that the compound image has the best visual effect. Then, change Eq.~\ref{allocation} to $\sigma(x_c) \propto 	{(\xi^*_{x_c}-\xi_{x_r})}/{(\xi_{w_r}-\xi_{x_r})}$. Simplify and approximate this equation to obtain:
\begin{equation}
	\label{contributor_simplify}
	\begin{split}
	\dfrac{1}{\sigma(x_c)} \propto  C^* - \dfrac{\xi^*_{x_c}-\xi_{w_r}}{\xi^*_{x_c}-\xi_{x_r}}
	\end{split}
\end{equation}
where $C^*$ is a constant. In conclusion, it can be inferred that both $\xi_{x_r}$ and $\xi_{w_r}$ determine the allocation strategy $\sigma(\cdot)$. Hence, the Nash equilibrium spot $(\xi^*_{x_r},\xi^*_{w_r})$ needs to be approximated by rewriting Eq.~\ref{game} on the T2IW scenario:
\begin{equation}
	\label{strategy_for_T2IW}
	\begin{split}		
		\sigma^{*}(x_c)&=\mathop{\mathrm{argmax}}\limits_{(s_{x_r},s_{w_r})} \mathrm{\Phi}_{gain}^{P}(s_{x_r},s_{w_r})\\
		&= \lim\limits_{\theta_x \to \theta_x^* \atop \theta_w \to \theta_w^*} \sigma(x_r,w_r;\theta_x,\theta_w)
	\end{split}
\end{equation}
where $s_{x_r}$ and $s_{w_r}$ are the strategies for decoupling the revealed image and the revealed watermark, respectively. $\theta_x$ and $\theta_w$ are the parameters of the decoupling networks for the image and the watermark, respectively, while $\theta_x^*$ and $\theta_w^*$ are the optimal parameters. It shows that the closer $(x_r,w_r)$ gets to the optimal solution $(x_r^*,w_r^*)$, the easier to enable Nash equilibrium. Note that $s_{x_r}$ and $s_{w_r}$ are forced to be executed to realize the trade-off without causing interference to one another.

Then, we have devised a decoupling method for the revealed image and the revealed watermark. For the revealed image, a considerable amount of information is expected to be retained comparable to the compound image, thereby reducing the disparity between them. An additional requirement is that the messages in the watermark should not be stored within the revealed image. In accordance with the Shannon mutual information theory~\cite{Russakoff04}, the procedure aims to shrink the difference between the revealed image and the compound image, whilst expanding the gap between the hidden watermark and the revealed image:
\begin{equation}
	\label{loss_revealed_image}
	\begin{split}
		&\mathop{\mathrm{argmax}}         \limits_{s_{x_r}}\mathrm{\Phi}_{gain}^{p_{x_r}}(s_{x_r}) \\
		&=\mathop{\mathrm{argmin}} \limits_{\theta_x} \mathcal{L}_{x_c \to x_r}(\theta_x)\\
		&=\mathop{\mathrm{argmin}}\limits_{\theta_x} [I_{M}(w_h,x_r;\theta_x)-{I_{M}}(x_c,x_r;\theta_x)]\\
		&s.t.\   I_{M}(x_c,x_r) \le I(x_c) -I(w_r)\\
	\end{split} 
\end{equation}
where $\mathrm{\Phi}_{gain}^{p_{x_r}}(s_{x_r})$ denotes the gains by executing the strategy $s_{x_r}$, $I(\cdot)$ represents the self-information function, $I_{M}(\cdot)$ represents the mutual information function to optimize the parameter $\theta_x$, calculated by the Kullback-Leibler divergence (KL divergence) $I_M(X, Y) = D_{KL}(P(X, Y) || P(X)P(Y))$. The less mutual information indicates the lower correlation between inputs. The goal of strategy $s_{x_r}$ is to minimize the objective $\mathcal{L}_{x_c \to x_r}(\theta_x)$ in order to extract the revealed image $x_r$. For the condition, the equal sign is only applicable when $x_r$ is ideal. To be constrained by Eq.~\ref{loss_revealed_image}, we first adopt a decoder to handle the compound image:
\begin{equation}
	\label{image_decoder}
	\begin{split}
		x_r = \mathbf{{R}}^x_\mathrm{U-Net}(x_c;{\theta_x})
	\end{split}
\end{equation}
where $ \mathbf{{R}}^x_\mathrm{U-Net}(\cdot)$ is the U-Net based revealed image decoder, establishing pixel-level dependencies between the input and the output. Diverging from the constraint stated in Eq.~\ref{loss_revealed_image}, the watermark requires measurements from both $x_r$ and $x_c$, see Fig.~\ref{fig:nash} (b). The idea situation is that the difference between mutual information $I_M(x_c,x_r)$ and self-information $I(x_c)$ represents the self-information of hidden watermark. As a result, $I(w_r)$ tries to search the space of the complementary set $\complement_{I(x_c)} I_M(x_c,x_r)$. The strategy and condition are described below:
\begin{equation}
	\label{loss_watermark}
	\begin{split}
		&\mathop{\mathrm{argmax}} \limits_{s_{w_r}}\mathrm{\Phi}_{gain}^{p_{w_r}}(s_{w_r}) \\
		&=\mathop{\mathrm{argmin}} \limits_{\theta_w} \mathcal{L}_{x_c \to w_r}(\theta_w) \\
		&=\mathop{\mathrm{argmin}} \limits_{\theta_w} \{[I(x_c;\theta_w) - I_M(x_c,x_r;\theta_w)]- I(w_r;\theta_w)\} \\
		&s.t.\  I(w_r) \textless I(x_c)\\
	\end{split}
\end{equation}
where the objective $\mathcal{L}_{x_c \to w_r}(\theta_w)$ forces the self-information of $w_r$ to stay away from the $I_M(x_c,x_r)$ space and approximate the $\complement_{I(x_c)} I_M(x_c,x_r)$ space. Furthermore, $I(w_r)$ has an upper bound $I(x_c)$. So the hidden information in $w_r$ can be recovered effectively. Also, similar to Eq.~\ref{image_decoder}, one decoder is needed to get the $w_r$:
\begin{equation}
	\label{watermark_decoding}
	\begin{split}
		w_r =  \mathbf{{R}}^w_\mathrm{U-Net}(x_c;\theta_w)
	\end{split}
\end{equation}
where the $  \mathbf{{R}}^w_\mathrm{U-Net}(\cdot)$ is the U-Net based revealed watermark decoder. Therefore, under the selfish strategies and decoders, the image and the watermark can be decoupled separately from a compound image towards the Nash equilibrium state. 

\subsection{Optimization Strategy}
As mentioned in Eq.~\ref{task_define}, we expect the T2IW framework to produce a compound image $x_c$ publicly and decouple the revealed image $x_r$ and the revealed watermark $w_r$ by specific decoders. For the compound image $x_c$, following the adversarial training scheme in~\cite{Tao00JBX22}, the objective function of the discriminator is defined as: 
\begin{equation}
	\label{loss_d_x_w}
	\begin{split}
		\mathcal{L}^D_{x_c}=&\mathbb{E}_{x\sim p_{r}} [log(D_{x_c}(x,s))]\\
		&+(1/2)\mathbb{E}_{x_c\sim p_{x_c}} [log(1-D_{x_c}(x_{c},s))]\\
		&+(1/2)\mathbb{E}_{x\sim p_{r}} [log(1-D_{x_c}(x,\hat{s}))]\\
	\end{split}
\end{equation}
where $\hat{s}$ corresponds to the description of the mismatched text, $p_r$ and $p_{x_c}$ denote the real data distribution and the compound image distribution, $D_{x_c}(\cdot)$ denotes the downsampling discriminant function. To ensure authentic expressiveness, the objective function of the generator for the compound image is improved as:
\begin{equation}
	\label{loss_g_x_w}
	\begin{split}
		\mathcal{L}^G_{x_c}= -\mathbb{E}_{x_c \sim p_{g}}[log (D_{x_c}(x_c,s))]\\		
	\end{split}
\end{equation}

Referring to the strategy in Eq.~\ref{loss_revealed_image}, the procedure for the revealed image $x_r$ can be strengthened toward:
\begin{equation}
	\label{loss_x_r}
	\begin{split}
		\mathcal L_{x_r} = \lambda_1 \cdot\mathcal{L}_{x_c \to x_r} +  ||x_r,x_c||_1
	\end{split}
\end{equation}
where $||\cdot||_1$ refers to the smooth-$L_1$ loss, and $\lambda_1$ is a proportional coefficient.
Eq.~\ref{loss_x_r} removes the watermark using $\mathcal{L}_{x_c \to x_r}$ while maintaining a similar visual appearance between the compound image and the revealed image through the smooth-$L_1$ loss. Apart from Eq.~\ref{loss_watermark}, we chase the mandatory constraint to make $w_r$ intact, and the objective function for recovering the watermark $w_r$ is as follows:
\begin{equation}
	\label{loss_w_r}
	\begin{split}
		\mathcal{L}_{w_r} = \lambda_2 \cdot\mathcal{L}_{x_c \to w_r} +  \delta(w_h,w_r)
	\end{split}
\end{equation}
where $\delta(\cdot)$ measures the similarity between the hidden watermark $w_h$ and the revealed watermark $w_r$ by using the MSE-$L_2$ loss, and $\lambda_2$ is a proportional coefficient. In total, we allow Eq.~\ref{loss_d_x_w} to conduct training first. Afterward, Eq.~\ref{loss_g_x_w}, Eq.~\ref{loss_x_r} and Eq.~\ref{loss_w_r} carry out gradient descent for $\mathbf{E_{\mathrm{U-Net}}}(\cdot)$, $  \mathbf{{R}}^x_\mathrm{U-Net}(\cdot)$, $ \mathbf{{R}}^w_\mathrm{U-Net}(\cdot)$ and other modules together.

\section{Evaluation Metrics}
\label{metrics}
Considering that the existing evaluation criteria for watermark generation and image generation are incomplete for the T2IW task, we design a full-scale evaluation scheme meticulously incorporating three distinct perspectives: image quality, watermark invisibility, and watermark robustness. 
\subsubsection{Image Quality}
To demonstrate that synthetic images are still photorealistic along with watermark perturbations, the Inception Score (IS)~\cite{SalimansGZCRCC16} and the Fréchet Inception Distance (FID)~\cite{HeuselRUNH17} are adopted. Both metrics serve for the compound images and the revealed images.

\textbf{IS} uses a pre-trained classifier Inception v3 network learned from ImageNet dataset~\cite{SalimansGZCRCC16} to measure the clarity and diversity of generated images, calculated as:
\begin{equation}
IS = exp(E_{x_g \sim P_g}D_{KL}(p(y|x_g)||p(y)))
\end{equation}
where the $x_g$ are the generated images and $y$ are the categories, $P_g$ is the generated data distribution. The KL divergence between the conditional distribution $p(y|x_g)$ and the marginal distribution $p(y)$ should increase to obtain a higher score.

\textbf{FID} computes the distance between the real data and the generated data distributions. Compared with the Inception net~\cite{SalimansGZCRCC16} in IS, the last full-connected layer is removed and the pooling vectors are extracted as outputs instead, by the following calculation:
\begin{equation}
	FID = \left|\left| \mu_r - \mu_g \right|\right|^2 + T_r(\Sigma r + \Sigma g) - (\Sigma  r \Sigma  g)^\frac{1}{2}
\end{equation}
where $\mu_r$ and $\mu_g$ are the mean feature of real images and generated images, $\Sigma r$ and $\Sigma g$ are covariance matrices, and $T_r$ is the trace of $\Sigma r$.

\subsubsection{Watermark Invisibility} The imperceptibility of hidden watermarks could be measured by widely-used metrics, i.e., Peak Signal-to-Noise Ratio (PSNR)~\cite{DingMCL22}, Structural Similarity (SSIM)~\cite{DingMCL22}, and Learned Perceptual Image Patch Similarity (LPIPS)~\cite{ZhangIESW18}.

\textbf{PSNR} reveals the invisibility of watermarks through pixel-level difference, which is essentially a image similarity evaluation indicator, calculated as:
\begin{equation}
	\label{psnr}\small
	\begin{split}
		&PSNR(x_c,x_r) = 10 \cdot log_{10}\Big(\frac{(2^d - 1)^2}{MSE}\Big)\\
		&MSE = \frac{1}{m \cdot n}\sum_{i=0}^{m-1} \sum_{j=0}^{n-1}||x^c_{i,j} - x^r_{i,j}||^2\\
	\end{split}
\end{equation}
where $MSE$ is the mean square deviation of the compound images and the revealed images, $d$ denotes the bit depth and is set to 8, $x^r_{i,j}$ and $x^c_{i,j}$ represent the pixels in the $i$-$th$ row and $j$-$th$ column of the revealed images and compound images. The unit of PSNR is $dB$.

\textbf{SSIM} is measured based on three tracks between the sample $x_c$ and the sample $x_r$: brightness $l$, contrast $c$, and structure $s$. The calculation process is represented as:
\begin{equation}
	\label{ssim}\small
	\begin{split}
		SSIM(x_c,x_r) &  = [l(x_c,x_r)]^\alpha \cdot [c(x_c,x_r)]^\beta \cdot [s(x_c,x_r)]^\gamma\\
		&l(x_c,x_r) = \frac{2\mu_c \mu_r+c_1}{\mu_c^2+\mu_r^2+c_1}\\
		&c(x_c,x_r) = \frac{2\sigma_c \sigma_r+c_2}{\sigma_c^2+\sigma_r^2+c_2}\\
		&s(x_c,x_r) = \frac{\sigma_{cr}+c_3}{\sigma_c \sigma_r +c_3}\\
	\end{split}
\end{equation}
where $\mu_c$, $\mu_r$, $\sigma_c$ and $\sigma_r$ denote the mean and variance of the compound images and the revealed images respectively, and $\sigma_{cr}$ is the covariance matrix. Empirically, $\alpha$, $\beta$ and $\gamma$ are set to 1. $c_1=(0.01L)^2$, $c_2=(0.03L)^2$ and $c_3=c_2/2$ are constants, where $L=2^d-1$. A higher SSIM score reflects the higher invisibility. 

\textbf{LPIPS} is more in line with human perception ability compared with traditional metrics~\cite{ZhangIESW18}. It forces the deep network to compute the reverse mapping of the revealed images from the compound images, and prioritizes the perceptual similarity between them. The calculation is as follows:
\begin{equation}\small
	LPIPS(x_c,x_r) = \sum_l \frac{1}{m_l \cdot n_l} \sum_{i=0}^{m-1} \sum_{j=0}^{n-1}||w_l \cdot (y^c_{i,j}-y^r_{i,j})||^2_2
\end{equation}
where the $y^c_{i,j}$ and $y^r_{i,j}$ are the compound image feature and the revealed image feature from the $l$-layer feature stack. The learned vector $w_l$ scales feature size and the $L_2$ distance $||\cdot||_2$ is ultimately calculated, which is expected to be lower to show that the watermarks are noticeable on the generated images.

\subsubsection{Watermark Robustness}
The watermarks fulfill the duty of carrying messages, although they encounter post-processing attacks. Therefore, the robustness evaluation scheme for watermarks needs to be constructed. Notice that the conventional metric Normalized Correlation (NC)~\cite{TancikMN20} is intractable to compute the character retention in the watermarks under attack. Nevertheless, NC can only unilaterally assess the spatial pixel-level similarity and overlook the character-level accuracy. Therefore, we develop a character reader using an effective Optical Character Recognition (OCR) tool~\cite{jeeva2022} and the edit distance string matching algorithm~\cite{MarzalV93} to solve this problem, called Character Accuracy (CA).
Both NC and CA can cooperatively measure the watermark robustness at the space-level and character-level.

\textbf{NC} is identical to the Euclidean distance and the Pearson correlation coefficient~\cite{DingMCL22}, it measures the correlation distance between the hidden watermarks and the revealed watermarks in statistics. The definition is as follows:
\begin{equation}
		\footnotesize{
		NC(w_r,w_h) =  \frac{ \sum_{i=0}^{m-1} \sum_{j=0}^{n-1} (w^r_{i,j}-\bar{w}_r)(w^h_{i,j}-\bar{w}_h)}{\sqrt{\sum_{i=0}^{m-1} \sum_{j=0}^{n-1} (w^r_{i,j}-\bar{w}_r)^2\sum_{i=0}^{m-1} \sum_{j=0}^{n-1}(w^h_{i,j}-\bar{w}_h)^2}}
	}
\end{equation}
where $w^h_{i,j}$ and $w^r_{i,j}$ are the pixel values of the hidden watermarks and the revealed watermarks at the $(i,j)$ coordinate, while $\bar{w}_r$ and $\bar{w}_h$ represent average RGB intensity. Within the range of -1 to 1, the larger $NC$, the higher the degree of restoration of the watermark, whether there are attacks or not.

\textbf{CA} is determined using faithful OCR and edit distance techniques to assess the correspondence of characters between hidden watermarks and revealed watermarks. The calculation is streamlined as follows:  
\begin{equation}
	CA(x_c,x_r) = \mathcal{D}_{edit}(OCR(x_c),OCR(x_r))
\end{equation}
where CA gets the character-level measurement for complementing the spatial similarity, $\mathcal{D}_{edit}(\cdot)$ is the edit distance computation, and $OCR(\cdot)$ is the OCR operation to extract characters from the watermarks. The results in Sec.~\ref{attacks} show the 
effectiveness of our proposed evaluation metric.

\section{Experiments}
In this section, we present the experimental results. It starts with the introduction of the experimental setup in Sec.~\ref{setup}. Then, quantitative evaluations are performed from three aspects: image quality in Sec.~\ref{img_quality}, watermark invisibility in Sec.~\ref{eva_Invisibility}, and watermark robustness in Sec.~\ref{attacks}. To demonstrate the effectiveness of modules in our proposed T2IW framework, an ablation study is provided in Sec.~\ref{ablation_study}. In addition, a parameter study on the coupling of features of the semantic feature and the watermark feature is given in Sec.~\ref{parameter_analysis}. Further visualizations in Sec.~\ref{visualization} intuitively show the superiority of image quality, watermark invisibility, and watermark robustness regarding our proposed T2IW.

\subsection{Experimental Setup}
\label{setup}
\subsubsection{Datasets}
To evaluate our proposed framework, we refer to the previous methods~\cite{Tao00JBX22,ye2023recurrent,ZhangS21,ZhangXL17} to verify the most popular benchmark T2I datasets: Oxford-102 flowers~\cite{NilsbackZ08}, CUB-birds~\cite{wah2011caltech} and MS-COCO~\cite{lin2014}. The concise introduction will be provided below.

\textbf{Oxford-102 Flower} consists of 102 categories of flowers with 8,189 images in total, while each class contains 40 $\sim$ 250 images of varying sizes, and each example contains 1 image with 5 captions. Similar to~\cite{reed2016}, we divide it into class-disjoint training split and test split.

\textbf{CUB-birds} involves 11,788 images, being assigned to 200 species of birds. Inherit from predecessors, we split them into 8,855 class-disjoint training images and 2,933 test images, which are associated with 10 manually labeled captions.

\textbf{MS-COCO} is a large-scale dataset with complex and ever changing scenarios. Statistically, 82,783 images are used for training, while 40,470 images are used for testing. Each image is described by five semantically similar captions for different perspectives of annotators.
\begin{center} 
\begin{table*}[!t]
	\footnotesize 
    \renewcommand\arraystretch{1.1}
	\centering
	\small
	\setlength{\tabcolsep}{5.4mm}{
		\caption{\small Evaluation about image quality of our framework applicable to two models on the three widely-used datasets.}
		\label{tb:image_quality}
		\vspace{-1ex}
		\begin{tabular}{ccccccc}
			\toprule
			\toprule
			\multirow{2}{*}{\textbf{Models}}
			&\multirow{2}{*}{\textbf{Mode}}
			&\multicolumn{2}{c}{\textbf{CUB-birds}}&\multicolumn{2}{c}{\textbf{Oxford-102 flowers}}&\multicolumn{1}{c}{\textbf{MS-COCO}}\\
			
			\cline{3-7}
			& &IS$\uparrow$ &FID$\downarrow$ &IS$\uparrow$ &FID$\downarrow$&FID$\downarrow$\\
			\midrule
			\multirow{3}{*}{RAT-GAN~\cite{ye2023recurrent}}
			&Raw Images  &$5.36$ $\pm$ $0.20$&$13.91$ &$4.09$ $\pm$ $0.06$ &$16.04$& $14.60$ \\
			
			&Compound Images &$ 4.94$ $\pm$ $0.06$ &$16.95$ &  $3.72$ $\pm$ $0.07$ &$18.35$& $16.32$\\
			&Revealed Images &$4.98$ $\pm$ $0.06$& $17.32 $& $ 3.81$ $\pm$ $0.07 $&$19.16$& $16.68$\\
			\midrule
			\multirow{3}{*}{AttnGAN~\cite{XuZHZGH018}}
			&Raw Images  & $4.36$ $\pm$ $0.03$&$23.98 $& -- & --&  $35.49$\\
			&Compound Images & $4.02$ $\pm$ $0.05$ &$26.49$&  -- &-- & $38.51$\\
			&Revealed Images & $4.09$ $\pm$ $0.05$ &$26.01$& -- &-- & $38.29$\\
			
			\bottomrule
			\bottomrule
		\end{tabular}
	}
	\vspace{-2ex}
\end{table*}
\end{center}
\begin{table*}[!t]
	\footnotesize 
	\renewcommand\arraystretch{1.5}
	\centering
	\setlength{\tabcolsep}{2.7mm}{
		\caption{\small Evaluation about watermark invisibility of our framework applicable to two models on the three widely-used datasets.}
		\label{tb:invisibility}
		\vspace{-1ex}
		\begin{tabular}{c|ccc|ccc|ccc}
			\toprule
			\toprule
			\multirow{2}{*}{\textbf{Models}}
			&\multicolumn{3}{c|}{\textbf{CUB-birds}}&\multicolumn{3}{c|}{\textbf{Oxford-102 flowers}}&\multicolumn{3}{c}{\textbf{MS-COCO}}\\
			
			\cline{2-10}
			&PSNR (dB)$\uparrow$ &SSIM(\%)$\uparrow$&LPIPS$\downarrow$ &PSNR (dB)$\uparrow$ &SSIM(\%)$\uparrow$&LPIPS$\downarrow $&PSNR (dB)$\uparrow$&SSIM(\%)$\uparrow$&LPIPS$\downarrow$\\
			\midrule
			\multirow{1}{*}{RAT-GAN~\cite{ye2023recurrent}}
			
			& 33.29 &98.46&0.0257& 33.51 &98.60& 0.0231&33.97& 98.77 &0.0219\\
			\midrule
			\multirow{1}{*}{AttnGAN~\cite{XuZHZGH018}}
			& 33.86 &98.15&0.0223&  -- &--&--&33.54 & 98.26&0.0235\\
			\bottomrule
			\bottomrule
		\end{tabular}
	}
	\vspace{-2ex}
\end{table*}
\subsubsection{Text-to-image Models} 
Referring to Sec.~\ref{related_work}, we choose GAN-based models from both single-stage and multi-stage generation as the classification category to respectively validate our proposed T2IW framework. Specifically, we select RAT-GAN~\cite{ye2023recurrent} and AttnGAN~\cite{XuZHZGH018} as they are widely known and have demonstrated good performance in T2I. These models are considered representative methods of two generative paradigms. Thus, we incorporate RAT-GAN and AttnGAN into our framework.

\subsubsection{Implementation Details}
The text encoder is frozen during training with an output of size 256. Random noise is sampled from a standard Gaussian distribution with 100 dimensions.  Adam optimizer is used to optimize the network with base learning rates of 1$\times$10$^{-4}$ for the generator and 4$\times$10$^{-4}$ for the discriminator. The kernel size in the up-sampling and down-sampling blocks is set to 3$\times$3. The binary watermark is set to single channel to minimize the impact on the expressiveness of the image with the size of 256$\times$256. $\lambda_1$ and $\lambda_2$ are set to 0.1. Empirically, we adopt a warm-up policy to synthesize the images without strategy $\mathrm{\Phi}_{gain}^{p_{w_r}}$ and strategy $\mathrm{\Phi}_{gain}^{p_{x_r}}$ for 50 epochs, preventing interference in the image generation manifold. All experiments are implemented on the NVIDIA RTX 3090 GPU computing platform.

\subsection{Evaluations about Image Quality}
\label{img_quality}
As the statement in Sec.~\ref{related_work}, we choose the widely known T2I models, that is, RAT-GAN~\cite{ye2023recurrent} and AttnGAN~\cite{XuZHZGH018}, from two generative paradigms, that is, single-stage generation~\cite{ye2023recurrent,Tao00JBX22,Tao_2023_CVPR} and multi-stage generation~\cite{ZhangXL17,ZhangXLZWHM19,XuZHZGH018}, to verify the generalization of the proposed T2IW method. Three modes of images are listed to be compared: \textit{1) raw images} are synthesized by the baseline models; \textit{2) compound images} refer to images with hidden watermarks, acquired from the generator of our framework; \textit{3) revealed images} are the unwatermarked version, reconstructed from the compound images. Fundamentally, we perform thorough experiments on three datasets and employ the well-known metrics (i.e., IS and FID) to evaluate the fidelity of the images. In the ideal case, the compound images, the revealed images, and the raw images should exhibit the almost near-identical appearance, despite with minor pixel-level disturbances. Reflected in the metrics, there should be slight performance fluctuations.

As the results on image quality are organized in Tab.~\ref{tb:image_quality}, we notice that our T2IW framework consistently achieves excellent performance on IS and FID, proving that the generated images are visually realistic even with embedding watermarking. To further substantiate this claim, we will discuss the quantitative results of two categories in Fig.~\ref{fig:quality}. 

\textbf{T2IW on RAT-GAN}, considered one of the representative \textit{single-stage} generation methods~\cite{ye2023recurrent,Tao00JBX22,Tao_2023_CVPR}, implements a slight attenuation on unitary-type datasets such as Oxford-102 flowers and CUB-birds, i.e. $0.37$ and $0.42$ fall on IS, $2.31$ and $3.04$ increase on FID about compound images, respectively. On the sample-diverse dataset, that is, MS-COCO, watermarks that carry messages produce a negligible affect on the compound image distribution, reflected by the increase $1.72$ in FID. For the revealed images, powered by the U-Net decoder, they can be recovered as excellently as raw images, where the sacrifice of IS and FID is subtle among the three datasets.

\textbf{T2IW on AttnGAN} inherits the \textit{multi-stage} skeleton from StackGAN~\cite{ZhangXL17} and StackGAN++~\cite{ZhangXLZWHM19}, driven by the attention mechanism~\cite{VaswaniSPUJGKP17} to focus on the salient regions. It adopts three stages to gradually refine the images. Here, we probe that the compound images that originated from our T2IW framework still maintain low quality degradation via multiple modifications. Note that the compound images and revealed images perform approximate to the raw images on IS and FID. Representatively, FID of compound images has increases $8.46\%$ and $8.51\%$ in the CUB-birds and MS-COCO datasets, and the hiding of the watermark will not hinder image synthesis. Worth mentioning, AttnGAN~\cite{XuZHZGH018} provided no official performance on Oxford-102 flowers.


In conclusion, the evaluations on synthetic image quality demonstrate that the watermark containing dense information has a negligible effect on the quality of the output, and our proposed T2IW framework has a satisfactory level of generalization for the single-stage and the multi-stage generative models. 

\begin{figure*}\centering
	\centering
	\includegraphics[scale=0.149]{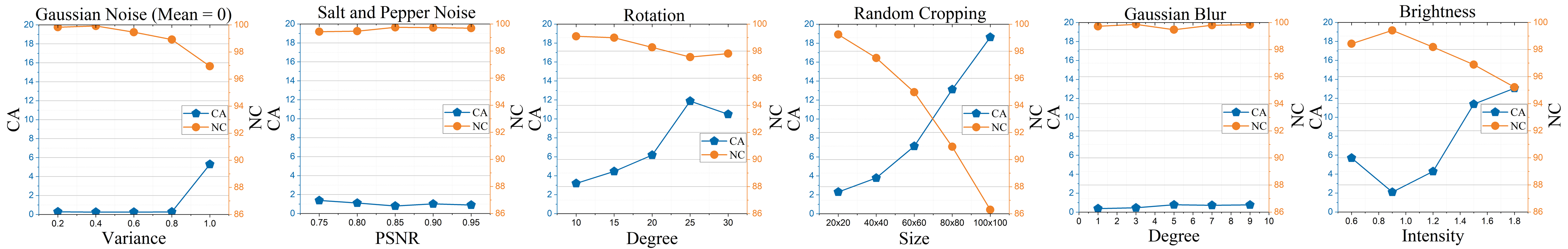}
	\vspace{-1ex}
	\caption{The curves for watermark robustness about T2IW on RAT-GAN under various post-processing attacks of different intensities, i.e, Gaussian noise, salt and pepper noise, rotation, random cropping, Gaussian blur and brightness. }
	\vspace{-1ex}
	\label{fig:attack}
\end{figure*}

\begin{table}[!t]
	\footnotesize 
	\renewcommand\arraystretch{1.3}
	\centering
	\setlength{\tabcolsep}{1.5mm}{
		\caption{\small Evaluation about watermark reconstruction of our framework applicable to two models on the three widely-used datasets.}
		\label{tb:nc_ca}
		\vspace{-1ex}
		\begin{tabular}{c|cc|cc|cc}
			\toprule
			\toprule
			\multirow{2}{*}{\textbf{Models}}
			&\multicolumn{2}{c|}{\textbf{CUB-birds}}&\multicolumn{2}{c|}{\textbf{Oxford-102 flowers}}&\multicolumn{2}{c}{\textbf{MS-COCO}}\\
			
			\cline{2-7}
			&NC(\%)$\uparrow$  &CA$\downarrow$  &NC(\%)$\uparrow$  & CA$\downarrow$  &NC(\%)$\uparrow$ & CA$\downarrow$ \\
			\midrule
			\multirow{1}{*}{RAT-GAN~\cite{ye2023recurrent}}
			&99.75 & 0.21 &99.69& 0.19 &99.81& 0.19\\
			\midrule
			\multirow{1}{*}{AttnGAN~\cite{XuZHZGH018}}
			& 99.72 &0.23 &--&-- & 99.48&0.21\\
			\bottomrule
			\bottomrule
		\end{tabular}
	}
	\vspace{-2ex}
\end{table}

\subsection{Evaluations about Watermark Invisibility}
\label{eva_Invisibility}
The invisibility of hidden watermarks should be reflected in the fact that they are imperceptible to the human visual system, when compatibility between textual semantics and watermark signals has been achieved. This requires that compound images do not produce explicit information leakage. Calculated as Eq.~\ref{psnr}, we adopt PSNR~\cite{DingMCL22} to measure the distinction between compound images and revealed images. As illustrated in Tab.~\ref{tb:invisibility}, the PSNR values of RAT-GAN and AttnGAN reach $33.29$dB and $33.86$dB in CUB-birds, respectively, while almost equal PSNR values are obtained in Oxford-102 flowers and MS-COCO. Referring to~\cite{MnihKSGAWR13}, the PSNR threshold is recommended as $30$dB, and our results obviously exceed this limit. Thus, the results of PSNR confirm the high imperceptibility of various watermarks in our method. 

Earlier work~\cite{DingMCL22} has pointed out that PSNR with a high value may not be fully compliant with the perceptive characteristic of the human visual system, of which SSIM~\cite{DingMCL22} takes special consideration instead. By imitating the human perceptual preference through the three perspective, computed as in Eq.~\ref{ssim}, we employ SSIM for further evaluation, the value of which ranges from $0\%$ $\sim$ $100\%$. As shown in Tab.~\ref{tb:invisibility}, the SSIM of two models maintain more than $98\%$ degree of matching, demonstrating the compatibility of the watermark and the semantic in pixels. Therefore, we obtain near-traceless information hiding in our T2IW framework. 

To focus on the intrinsic structure of the feature of the generated images, the LPIPS model~\cite{ZhangIESW18} is used to learn the perceptual distance of the compound images and the revealed images, as shown in Tab.~\ref{tb:invisibility}.  RAT-GAN and AttnGAN produce plausible values, that is, $0.0219$ and $0.0235$ on MS-COCO, which are lower than the existing method~\cite{LuNSX22} that reached $0.0320$ in the case of watermarking real-world images. In other words, the watermarks hidden in the synthesis images can hardly be perceived. 

Henceforth, we provide the remarkable T2IW baseline performances for watermark invisibility. Whether by spatial pixels or feature perception, the compound images keep a minimal gap to the revealed images, confirming that the stunning invisibility of the revealed watermarks has been reaped. 

\begin{figure}\centering
	\centering
	\includegraphics[scale=0.42]{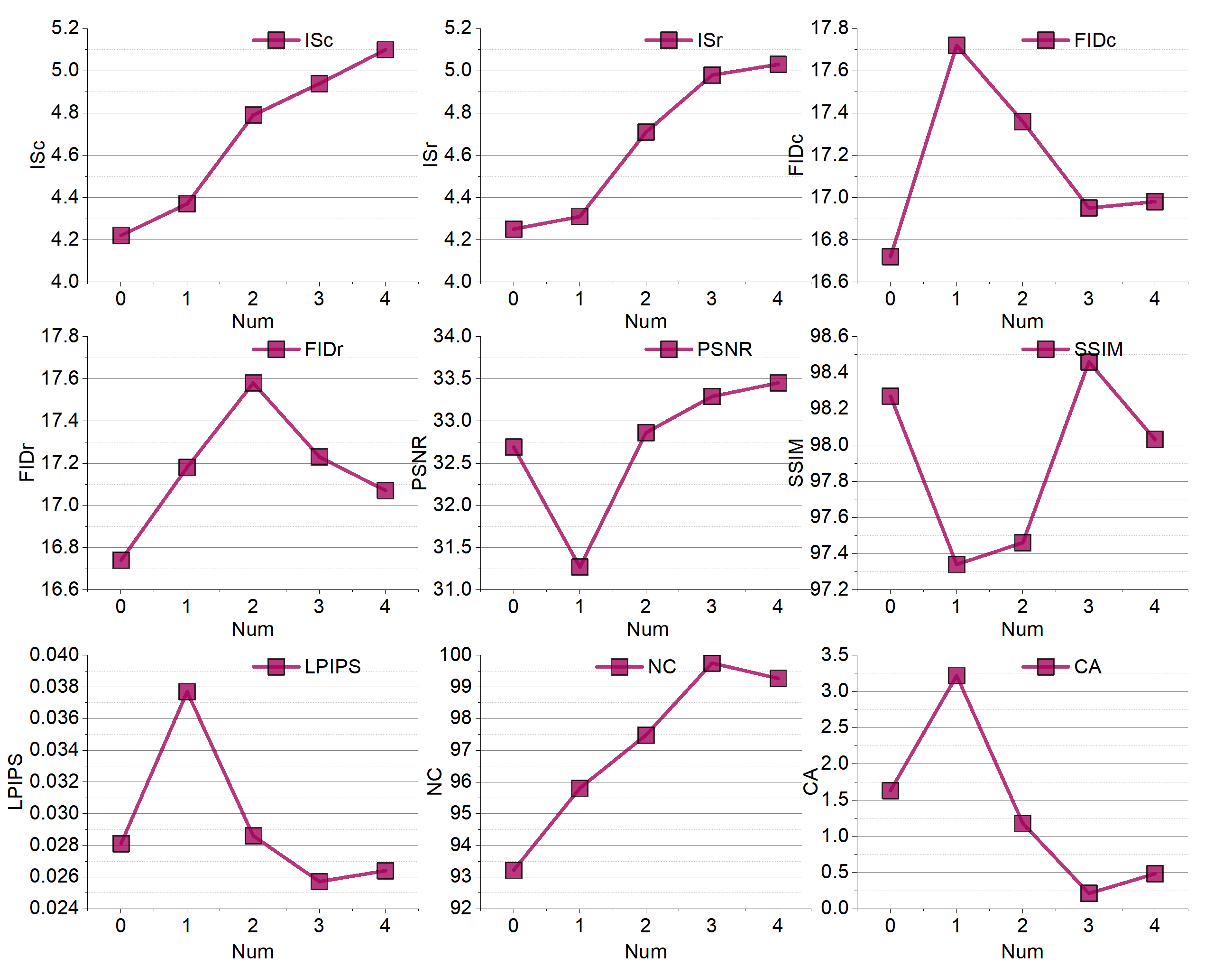}
	\vspace{-1ex}
	\caption{Curves of parameter analysis about the feature coupling
iteration number.}
	\vspace{-1ex}
	\label{fig:ablations}
\end{figure}

\subsection{Evaluations about Watermark Robustness}
\label{attacks}
We have conducted verification on the quality of watermark reconstruction from compound images, focusing particularly on the watermark reconstruction after undergoing post-processing attacks. Comprehensively, we list the results of watermark reconstruction in the presence and absence of attacks to evaluate the robustness of the watermark.
\subsubsection{Watermark Reconstruction without Attacks}
It should be noted that the T2IW framework is expected to reconstruct the non-destructive watermarks. Here, the quality of revealed watermarks is directly contrasted with the original hidden watermarks without attacks, as shown in Tab.~\ref{tb:nc_ca}. NC~\cite{DingMCL22} can contrapuntally measure spatial similarity in a pixel-by-pixel manner, showing that the similarity between revealed watermarks and hidden watermarks is greater than $99\%$ with few distorted pixels. Therefore, we achieve a high-level watermark reconstruction from a spatial perspective.

However, the pixels occupied by characters may be sparse, so the presence of information content can not be well evaluated spatially. Additionally, we employ the proposed CA, which combines OCR~\cite{jeeva2022} and edit distance~\cite{MarzalV93} to measure the character accuracy. Among the datasets and the models, the average CA is below $0.17$. It indicates that there are almost negligible character mismatches between the original and revealed strings. In other words, every character can be restored without being attacked.

\begin{table*}[t]
	\renewcommand\arraystretch{1.2}
	\centering
	\setlength{\tabcolsep}{0.8mm}{
		\caption{\small Ablation study about the T2IW on RAT-GAN for CUB-Birds dataset. ($\checkmark$) indicates ``applied''. (--) indicates ``removed''. }
		\label{tb:ablations}
		\vspace{-1ex}
		\begin{tabular}{ccccc|ccccccccc}
			\toprule
			\toprule
			\multirow{2}{*}{\textbf{U-Net Enc.}}
			&\multirow{2}{*}{\textbf{U-Net Dec.}}
			&\multirow{2}{*}{\textbf{Block $\mathbf{Y}_i$}}
			&\multirow{2}{*}{\textbf{Strategy $\mathrm{\Phi}_{gain}^{p_{w_r}}$}}
			&\multirow{2}{*}{\textbf{Strategy $\mathrm{\Phi}_{gain}^{p_{x_r}}$}}
			&\multicolumn{9}{c}{Metrics}\\
			\cline{6-14}
			&&&&&IS$_c$$\uparrow$&IS$_r$$\uparrow$&FID$_c$$\downarrow$&FID$_r$$\downarrow$&PSNR$\uparrow$&SSIM$\uparrow$&LPIPS$\downarrow$&NC$\uparrow$&CA $\downarrow$\\
			\midrule
			--&\checkmark&\checkmark&\checkmark&\checkmark  &$4.58\pm0.07$	&$4.64\pm0.07$	&$18.44$&$18.02$&$32.79$ &$96.03$	&$0.0301$ &	$99.02$&$1.28$\\ 
			\checkmark&--&\checkmark&\checkmark&\checkmark  &$4.75\pm0.06$ 	&$4.37\pm0.08$	&$17.33$ &$18.70$& $33.04$&	$97.72$&$0.0283$ &$98.26$ &$0.97$\\
			\checkmark&\checkmark&--&\checkmark&\checkmark &$4.22\pm0.06$&$4.25\pm0.06$	&$16.72$	&$\textbf{16.74}$	&$32.69$	&$98.27$&	$0.0381$&$93.21$&$6.63$\\
			\checkmark&\checkmark&\checkmark&--&-- &$\textbf{5.08}\boldsymbol{\pm}\textbf{0.05}$ &$4.89\pm0.05$	&$\textbf{16.61}$&$17.44$ &$33.11$	&$98.18$&$0.0294$ &$98.51$&$1.63$ \\	
			\checkmark&\checkmark&\checkmark&\checkmark&\checkmark&$4.94 \pm 0.06$	&$\textbf{4.98}\boldsymbol{\pm}\textbf{0.06}$&$16.95$&$17.32$&$\textbf{33.29}$	&$\textbf{98.46}$&$\textbf{0.0257}$&$\textbf{99.75}$&$\textbf{0.21}$\\
			\bottomrule
			\bottomrule
		\end{tabular}
	}
	\vspace{-2ex}
\end{table*}

\subsubsection{Watermark Reconstruction under Attacks}
In the real-world scenario where post-processing attacks such as Gaussian noise, rotation, random cropping, etc. may occur, the watermark requires to carry efficient robustness. Following the classic settings~\cite{ZhangLJYXZ19,MohanarathinamS20}, we invoke 6 attack modes with various intensities, plotted as curves about T2IW on RAT-GAN in Fig.~\ref{fig:attack}. In practice, we conduct attacks during training to make our decoder more adaptable to attack modes. The orange curves represent the results of NC, while the blue curves represent the results of CA. In most cases, there is an inverse relationship between the intensity of attacks and the preservation of watermark integrity, and CA provides a more accurate evaluation of the amount of information retained. 

We conclude the fluctuations of NC and CA here. \textit{1) Gaussian Noise:} When the mean value of the Gaussian noise distribution is set to $0$, the satisfying result can still be obtained even with the variance increasing to $1.0$, which demonstrates excellent resistance. \textit{2) Salt and Pepper Noise:} A higher PSNR represents less disturbance. NC exceeds $99\%$ from $0.75$ to $0.95$, showing that the watermark is almost unaffected. \textit{3) Rotation:} As the degree of rotation increases, the watermark exhibits more deterioration and is resistant to rotations of minor angles. \textit{4) Random Cropping:} Once the region is cropped, the watermark messages on the corresponding area will be lost. Therefore, a larger cropping area means more damage to the watermarks. \textit{5) Gaussian Blur:} In general, the impact on the watermark caused by this attack is minimal, and the watermark demonstrates strong resistance against such attacks. \textit{6) Brightness:} The more closely the attacked brightness returns to its original value (i.e., $1.0$), the more effectively the watermark will be restored.

In summary, our T2IW framework demonstrates significant robustness against post-processing attacks and effectively maintains the confidentiality of secret messages. Intuitively, the results of the visualization are presented in Sec.~\ref{visualization}, where we will elaborate in subsequent discussions.

\begin{figure*}\centering
	\centering
	\includegraphics[scale=0.615]{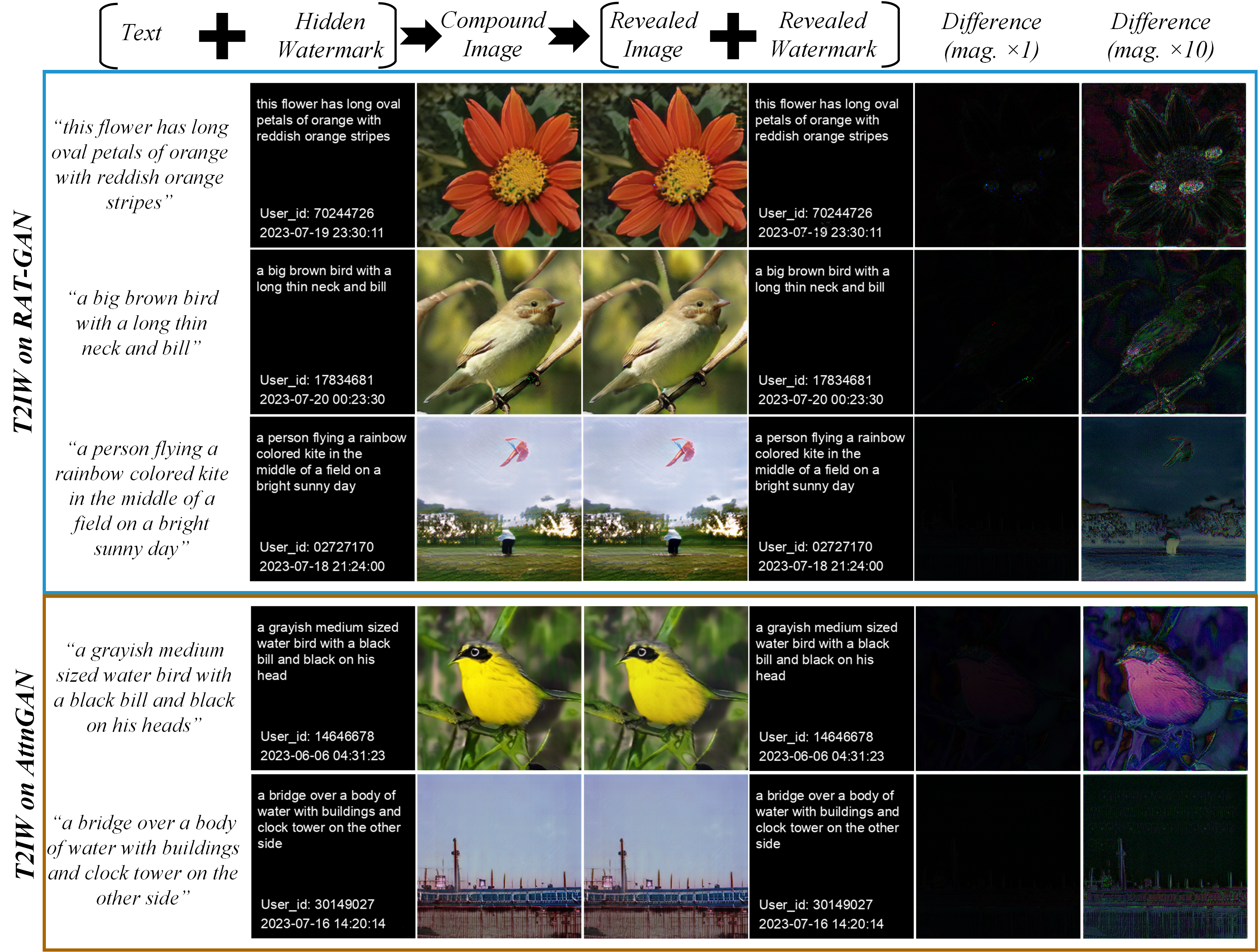}
	\vspace{-1ex}
	\caption{Overall visualizations from our T2IW framework applicable to two models on the three widely-used datasets. Difference images are obtained by subtracting pixels of the revealed image from pixels of the compound image, with magnifications of one and ten times for the pixel difference values.}
	\vspace{-1ex}
	\label{fig:quality}
\end{figure*}

\subsection{Ablation Study}
\label{ablation_study}
In order to showcase the efficacy of the modules integrated within our T2IW framework, we evaluate the important components by performing the ablation study on T2IW in RAT-GAN, as shown in Tab.~\ref{tb:ablations}. IS$_c$ and FID$_c$ are calculated by the compound images, while IS$_r$ and FID$_r$ are calculated by the revealed images. \textit{1) U-Net Enc.:} After removing the U-Net encoder, joint generation output images suffer due to weakening of the semantic and watermark feature dependency. \textit{2) U-Net Dec.:} The absence of the U-Net decoder results in the decoupling of the compromised revealed watermark and the revealed image, causing significant damage to IS$_r$, FID$_r$, NC and CA. \textit{3) Block $\mathbf{Y}_i$:} Block $\mathbf{Y}_i$ can make the fusion of semantic and watermark features more compatible to obtain superior IS$_c$ and FID$_c$. \textit{4) Strategies $\mathrm{\Phi}_{gain}^{p_{w_r}}$ and $\mathrm{\Phi}_{gain}^{p_{x_r}}$: } Both strategies serve the image decoupling and force the revealed image and the revealed watermark to optimize toward Nash equilibrium, resulting in a promotion of metrics such as PSNR, SSIM, LPIPS, NC, and CA.

\subsection{Parameter Analysis}
\label{parameter_analysis}
To harmonize the textual semantic and the watermark signal throughout the generation process, we probe the analysis for the feature coupling iteration number, i.e., the number of that $\mathbf{Y}_i$ is used, as illustrated in Fig.~\ref{fig:ablations} about T2IW on RAT-GAN.A higher fusion frequency will make the semantic feature more fully involved, thus reducing the proportion of the watermark signal. With increasing numbers, a higher IS indicates improved compatibility between the semantic feature and the watermark feature in the compound image. Additionally, the competitive relationship between the revealed watermark and the revealed image is mitigated. In summary, setting the feature coupling iteration number to three achieves an optimal balance for all metrics.

\begin{figure*}\centering
	\centering
	\includegraphics[scale=0.245]{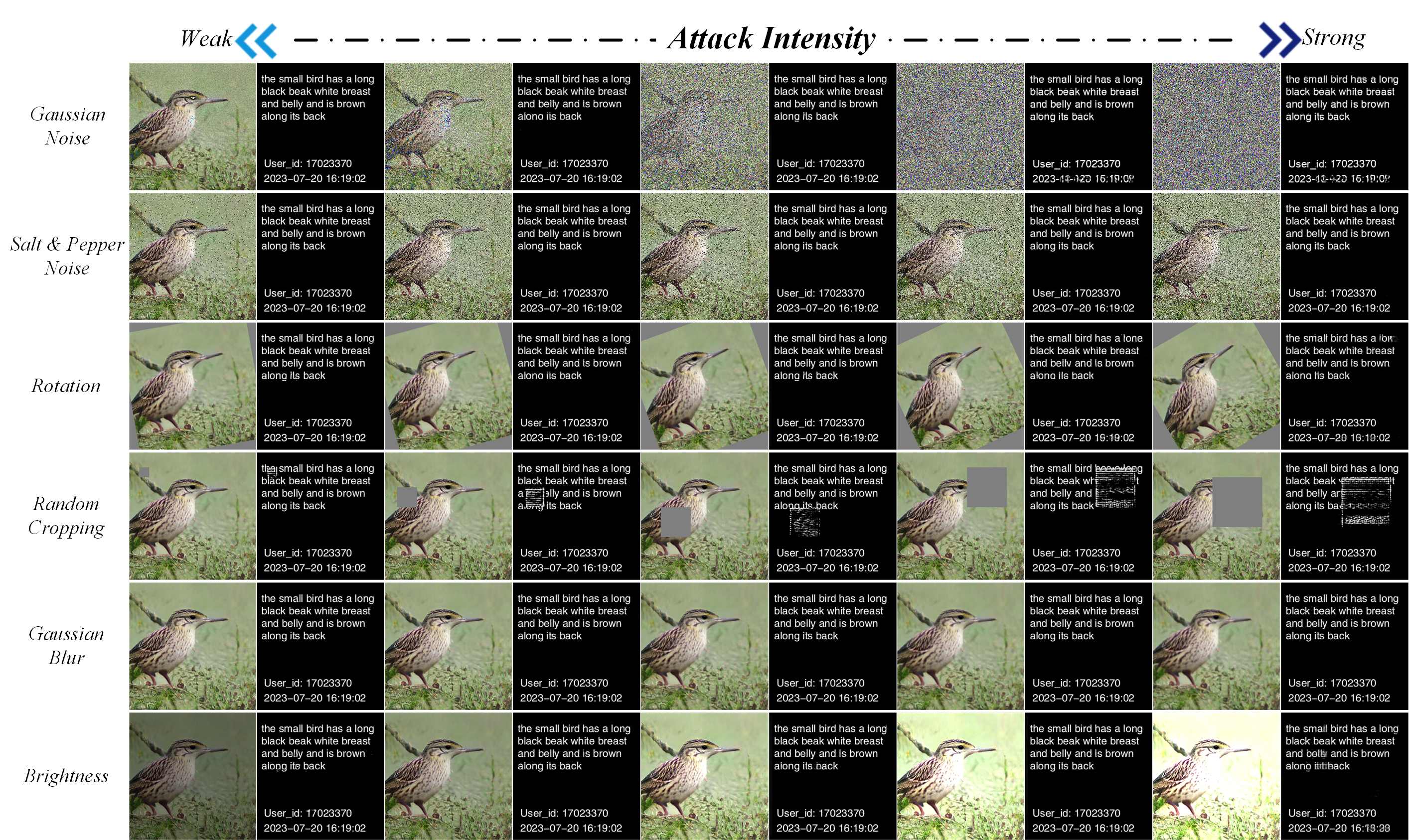}
	\vspace{-1ex}
	\caption{Visualizations of the watermark robustness under post-processing attacks with various intensities.}
	\vspace{-1ex}
	\label{fig:intensity}
\end{figure*}

\subsection{Visualization} 
\label{visualization}
In this section, we present a comprehensive display of visual outcomes on Oxford-102 flowers, CUB-birds and MS-COCO datasets about our T2IW framework on RAT-GAN. 
\subsubsection{Overall Qualitative Findings}The Fig.~\ref{fig:quality} showcases five instances, each comprising the compound image, the revealed image, the hidden watermark, the reconstructed watermark, the one- and ten-fold magnification differences. Interestingly, the presence of hidden watermark in the image does not significantly alter the visual appearance of the pattern, which exhibits subtle variations in chromatic without message leakage. Moreover, the revealed watermark enables a near-lossless reconstruction, and all characters are clearly visible. We have successfully achieved a remarkable visual effect, ensuring the trade-off of information allocation between the image and the watermark.

\subsubsection{Pixel-wise Difference} We guarantee that the revealed image and the compound image exhibit minimal pixel distance discrepancies, drawn in Fig.~\ref{fig:quality}. Non-zero pixels are rarely distributed in one-fold magnification differential maps, showing that the hidden watermark has been integrated into the image feature entirely. To highlight the differences more explicitly, we manually magnify the pixel errors by 10 times. One can observe that the secret messages are still not identifiable. Overall, these difference images demonstrate that our T2IW framework ensures to hide the watermark with only slight distortion of the image.


\begin{figure}\centering
	\centering
	\includegraphics[scale=0.42]{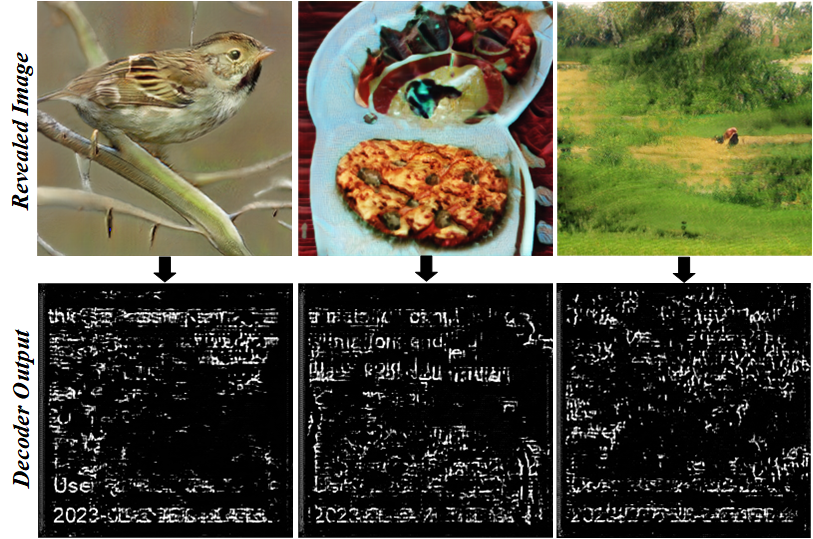}
	\vspace{-1ex}
	\caption{ Examples of extracting watermarks from the revealed images by the U-Net watermark decoder, which effectively demonstrates that the revealed images lack the capability to recover watermark signals.}
	\vspace{-1ex}
	\label{fig:revealed_img_extraction}
\end{figure}

\subsubsection{Attacking on Watermarks} 
To tackle the challenges of watermark reconstruction in real-world scenarios, we have introduced the integration of post-processing attacks in the training phase. This approach can deploy robust decoder parameters that are better equipped to handle various post-processing attacks, as illustrated in Fig.~\ref{fig:intensity}. We summarize the outcomes as follows:
\begin{itemize}
	\item When dealing with Gaussian noise and salt-and-pepper noise, the use of powerful convolutional kernels can effectively filter out watermark information within the receptive fields. Even in cases where the noise completely covers the pixels, the extracted watermark still retains a certain level of distinguishability.
	\item Higher degrees of rotation result in the erasure of more message content.
	\item Random cropping unavoidably corrupts the corresponding region of the watermark, while the remaining area remains intact and visible.
	\item The application of Gaussian blur reduces the high-frequency details in the image, yet the watermark is relatively unaffected.
	\item Adjusting the brightness of the image causes the watermark to weaken as it deviates from its original brightness. 
\end{itemize}
In general, the watermarks from the visual illustration ensure desirable identifiability. Our T2IW strives to preserve the quality of reconstruction of the watermark, ensuring its exceptional robustness under various post-processing attacks. 

\subsubsection{Visual Effect of Decoding Revealed Images} 
T2IW forces the revealed image away from carrying watermark information, as some instances are provided in Fig.~\ref{fig:revealed_img_extraction}. The characters in the extracted watermark are erased to be almost unrecognizable, where the white and black pixels are displayed due to the pixel-level binary output of the convolutional kernel in the decoder. Hence, the information in the revealed images is inadequate to reconstruct the watermarks.

\begin{figure}\centering
	\centering
	\includegraphics[scale=0.42]{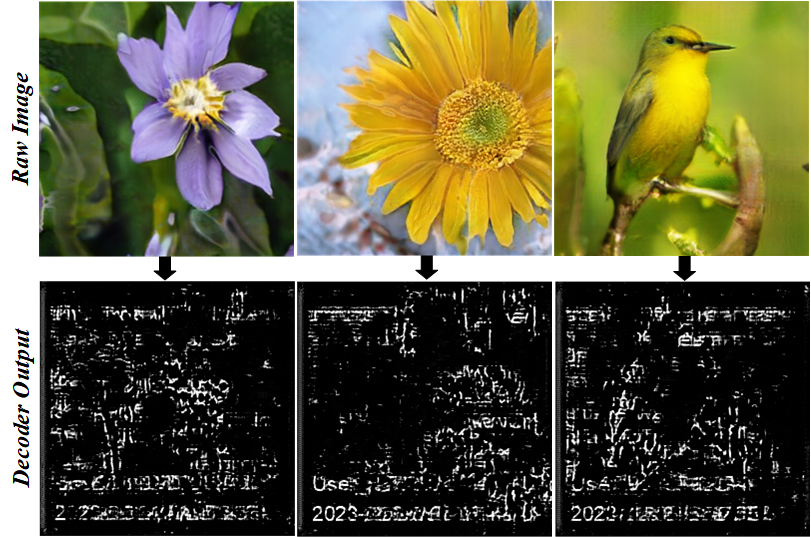}
	\vspace{-1ex}
	\caption{ Examples of extracting watermarks from the raw images by the U-Net watermark decoder. Essentially, the raw images produced by RAT-GAN do not inherently contain any watermark information and our decoder does not forge watermark data based on the learned patterns.}
	\vspace{-1ex}
	\label{fig:raw_img_extraction}
\end{figure}

\subsubsection{Visual Effect of Decoding Raw Images}
We attempt to use the decoder to extract messages from the raw images produced by RAT-GAN, as shown in Fig.~\ref{fig:raw_img_extraction}. Similar to Fig.~\ref{fig:revealed_img_extraction}, fuzzy characters such as "U", "2", and "0" appear due to the decoder's bias. Overall, our decoder cannot fabricate watermarks for the raw images based on the learned patterns.

\begin{figure}\centering
	\centering
	\includegraphics[scale=0.42]{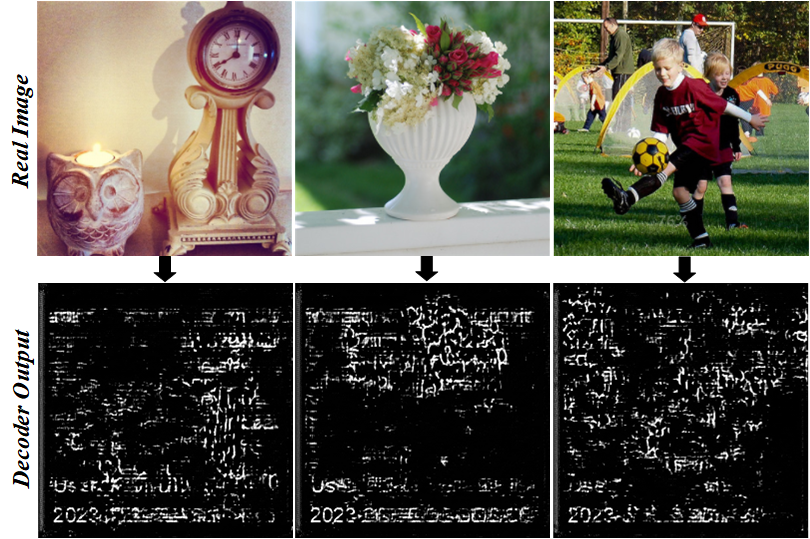}
	\vspace{-1ex}
	\caption{ Examples of extracting watermarks from the real images in COCO datasets by the U-Net watermark decoder. It confirms that our decoder  does not function properly on the watermark-free images.}
	\vspace{-1ex}
	\label{fig:real_img_extraction}
\end{figure}

\subsubsection{Visual Effect of Decoding Real Images}
Here, some images from the MS-COCO dataset are decoded by our watermark decoder, as shown in Fig.~\ref{fig:real_img_extraction}. The outputs of the decoder demonstrate the distorted visual appearance, further indicating that the watermark decoder is ineffective when applied to watermark-free images.

\section{Conclusion}
In this study, we introduce a new task called T2IW to address the issue of traceability for generated images. Correspondingly, we present a framework for T2IW that involves embedding the watermark signal into the generated image in an invisible manner. This approach can be demonstrated to generalize to single-stage and multi-stage GAN-based T2I models. Further, Shannon information theory and non-cooperative game theory are improved to effectively decouple the revealed watermark and the revealed image from a compound image. In order to resist common attacks of different intensities, we employ a data augmentation strategy that applies post-processing attacks to the compound image in the training phase. This strategy helps us to obtain the robust decoder parameters, resulting in strong robustness of the hidden watermark. In addition, we develop a customized evaluation scheme to evaluate the performance of our framework from the perspectives of image quality, watermark invisibility, and watermark robustness. Overall, through extensive experiments, we have demonstrated that our proposed framework generates compound images with outstanding visual effects and successfully reveals messages in the hidden watermarks even subjected to attacks. 


\bibliography{ref}{}
\bibliographystyle{IEEEtran}

\end{document}